\documentclass[10pt, a4paper]{article}
\usepackage{lrec2022} 
\usepackage{multibib}
\newcites{languageresource}{Language Resources}
\usepackage{graphicx}
\usepackage{tabularx}
\usepackage{soul}

\usepackage{xstring}
\usepackage{xcolor}
\usepackage{tikz-dependency}
\usepackage{adjustbox}
\usepackage{comment}
\usepackage[nobiblatex]{xurl}
\usepackage{xurl}
\usepackage{hyperref}       
\usepackage[capitalise,noabbrev]{cleveref}
\usepackage{enumitem}
\usepackage{svg}

\usepackage{titlesec}
\titleformat{\section}{\normalfont\large\bfseries\center}{\thesection.}{1em}{}
\titleformat{\subsection}{\normalfont\SmallTitleFont\bfseries\raggedright}{\thesubsection.}{1em}{}
\titleformat{\subsubsection}{\normalfont\normalsize\bfseries\raggedright}{\thesubsubsection.}{1em}{}
\renewcommand\thesection{\arabic{section}}
\renewcommand\thesubsection{\thesection.\arabic{subsection}}
\renewcommand\thesubsubsection{\thesubsection.\arabic{subsubsection}}

\usepackage{epstopdf}
\usepackage[utf8]{inputenc}

\usepackage{hyperref}
\usepackage{xstring}

\usepackage{color}

\title{Universal Dependency Treebank for Odia Language}

\name{Shantipriya Parida$^{1}$, Kalyanamalini Sahoo$^{2}$, Atul Kr. Ojha$^{3}$, \\ { \bf \large Saraswati Sahoo$^{4}$, Satya Ranjan Dash$^{5}$ and Bijayalaxmi Dash$^{6}$}
} 

\address{$^{1}$Silo AI, Helsinki, Finland \\
$^{2}$University of Lille, France \\ 
$^{3}$Insight Centre for Data Analytics, DSI, NUI, Galway, Ireland \\
$^{4}$Institute of Mathematics and Applications, India \\
$^{5}$KIIT University, Bhubaneswar, India \\
$^{6}$Ravenshaw University, Cuttack, India \\
shantipriya.parida@silo.ai, kalyanamalini.shabadi@univ-lille.fr, atulkumar.ojha@insight-centre.org, \\ sahoosaraswati455@gmail.com, sdashfca@kiit.ac.in, rudrabijayalaxmi@gmail.com\\
}

\abstract{
This paper presents the first publicly available treebank of Odia, a morphologically rich low resource Indian language. The treebank contains approx. 1082 tokens (100 sentences) in Odia selected from ``Samantar", the largest available parallel corpora collection for Indic languages. All the selected sentences are manually annotated following the ``Universal Dependency (UD)" guidelines. The morphological analysis of the Odia treebank was performed using machine learning techniques.
The Odia annotated treebank will enrich the Odia language resource and will help in building language technology tools for cross-lingual learning and typological research. We also build a preliminary Odia parser using a machine learning approach. The accuracy of the parser is 86.6\% Tokenization, 64.1\% UPOS, 63.78\% XPOS, 42.04\% UAS and 21.34\% LAS. Finally, the paper briefly discusses the linguistic analysis of the Odia UD treebank.
 \\ \newline  }

\begin{document}

\maketitleabstract

\section{Introduction}
\label{sect:intro}

Odia (earlier known as Oriya) is an Indian language belonging to the Indo-Aryan branch of the Indo-European language family. It is the predominant language of the Indian state of Odisha. Odia is written in Odia script, which is a Brahmic script. There are 37 million Odia speakers in India.\footnote{\url{https://censusindia.gov.in/2011Census/Language_MTs.html}} Odia is one of the many official languages of India and is designated as a Classical language.

Odia is an agglutinative language \cite{sahoo2001oriya}, and hence, a morphologically rich language. Odia’s verb morphology is rich with a three-tier tense system, person, number, and honorific markers. The prototypical word order is subject-object-verb (SOV) \cite{odiencorp1,odiencorp2}. Odia nominal morphology differentiates between plural and singular numbers; case marking on nouns; first, second, and third-person pronouns. But it does not have grammatical gender marking, which reduces the complexities of learning the language. Odia language allows Noun-verb, Adjective-verb, and Verb-verb compounding but does not allow elision. It has 28 consonants, 6 vowels, 9 diphthongs, and 4 semivowel phonemes. Most vowels can be short or long, and care must be taken to remember that the length of the vowel changes the word meaning completely. Odia's vocabulary is influenced by Sanskrit and also a little influence from Arabic, Persian, and Austronesian languages as the Kalinga empire (Odisha's ancient name) was connected to different other kingdoms.\footnote{\url{https://www.nriol.com/indian-languages/oriya-page.asp}} Odia language lacks online content and resources for natural language processing (NLP) research.

Unlike Treebanks of widely accepted languages such as English, Mandarin, Hindi, and Spanish for Natural Language Processing applications, applications based on low resource language like Odia is stagnated due to low resources. This paper is one step toward providing resources for such a low resource language. To start with we have worked on making a treebank in the Odia language. This project will surely help the Odia community and NLP researchers in providing resources for NLP applications.

\section{Odia Language Grammar}
\label{sect:odia_grammar}

Odia is an SOV language.  Usually, a simple sentence begins with a subject and ends with a finite verb. The major word classes found in Odia are nouns, pronouns, verbs, adjectives, and postpositions.  Certain minor categories like classifiers, complementizers, and conjunctions are also found. The objects occur between the subject and the verb, the Indirect Object precedes the Direct Object. The modifier precedes the item it modifies: the adjective precedes the substantive it qualifies, and the adverb precedes the verb. Although scrambling is allowed, usually, the word-order sticks to the V-final constructions except for poetic inversion \cite{sahoo2001oriya}. 

\paragraph{Declension}

Odia has two numbers: singular and plural; and three persons: 1st person, 2nd person, and 3rd person. The subject NP agrees with the verb in person, number, and honorific.
Honorificity goes along with person and number and it is marked in various word classes like nouns, pronouns, verbs, and, interestingly enough, with some of the post-positions that function as genitive, locative, and ablative markers. Generally, the person-number suffixes also go together. 

There are eight cases in Odia: nominative, accusative, instrumental, dative,  ablative, genitive, locative, and vocative.  Except for the nominative case, all the other cases are marked morphologically.

Phonologically, there is no distinction in the form of a word in masculine, feminine, or neuter gender in Odia.  E.g. \textit{pua} ‘son’ (masc), \textit{jhia} ‘daughter’ (fem), \textit{phaLa} ‘fruit’ (neuter).  But there are certain cases, where one finds such differences between the masculine and the feminine form of the words phonologically.  E.g. \textit{chhaatra} ‘male student’, \textit{chhaatri} ‘female student’. 

\paragraph{Pronouns}

\begin{figure}[ht](Sahoo, 2001)
    \begin{center}
    \includegraphics[scale=0.27]{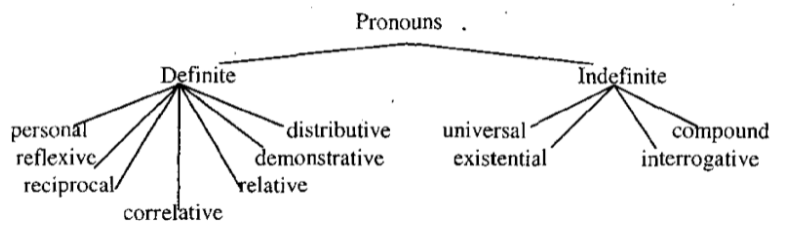}
    \caption{Odia Pronouns
    }
    \label{fig:odia_pronouns}
    \end{center}
\end{figure}

Odia pronouns are shown in ~\Cref{fig:odia_pronouns}.

\begin{itemize}[noitemsep]
    \item Personal: \textit{mun} ‘I’, \textit{tu} ‘you’, \textit{tume} ‘you’, \textit{aapaNa} ‘you’, \textit{se} ‘she’ / ‘he’
    \item Reflexive: \textit{se nije} ‘he himself / she herself’ 
    \item Reciprocal: \textit{paraspara} ‘each other’
    \item Correlative: \textit{jie} ‘who (ever)’ \textemdash \textit{se} ‘he’ / ‘she’
    \item Relative: \textit{je} ‘whoever’, \textit{jaahaaku} ‘whomsoever’ 
    \item Demonstrative: \textit{eha / ehi} ‘this’, \textit{eguDika / eguDaaka} ‘these’, sehi ‘that’ and seguDika / seguDaaka ‘those’
    \item Distributive: \textit{pratyeka} ‘each’ / ‘every’
    \item Universal: \textit{samaste} ‘all’
    \item Existential: \textit{jaNe} ‘one person’, \textit{goTe}  ‘a’/ ‘one’
    \item Interrogative: \textit{kie} ‘who’, \textit{kaahaaku} ‘whom’,’
    \item Compound: \textit{kehi jaNe} / \textit{kie jaNe} ‘somebody’
\end{itemize}

\paragraph{Case morphemes}

Odia case morphemes are shown in~\Cref{tab:odia_case}.

\begin{table}[ht]
    \centering
    \small
    \begin{tabular}{|p{2cm}|p{2cm}|p{2.5cm}|}
        \hline
        \textbf{Case} & \textbf{Singular} & \textbf{Plural/[+Hon] sg} \\
        \hline
        Nominative (NOM) & - & -e \\ \hline
        Accusative (ACC) & \textit{ku} & \textit{nku, maananku} \\ \hline
        Instrumental (INST) & \textit{re, dwaaraa, dei} & \textit{re, dwaaraa, dei} \\ \hline
        Dative (DAT) & \textit{ku} & \textit{nku, maananku} \\ \hline
        Ablative (ABL) & \textit{ru, Thaaru} & \textit{MaanankaThaaru} \\ \hline
        Genitive (GEN) & \textit{ra} & \textit{nkara, maanankara} \\ \hline
        Locative (LOC) & \textit{re, Thaare} & \textit{MaanankaThaare} \\ \hline
        Vocative (VOC) & \textit{he, bho} & \\ 
     \hline
    \end{tabular}
    \caption{The Case morphemes in Odia}
    \label{tab:odia_case}
\end{table}

\paragraph{Postpositional words}

The following postpositional words are used to express different case relations.

\begin{itemize}[noitemsep]
    \item \textit{aagare}  ‘before’
    \item \textit{pare} ‘after’
    \item \textit{kari}  ‘by’
    \item \textit{nimitte} ‘for
    \item \textit{parjyante} ‘up to’
    \item \textit{paain} ‘for’
    \item \textit{prati} ‘to’, ‘against’
    \item \textit{baahaara} ‘out’, ‘outside’
    \item \textit{byatita} ‘without’
    \item \textit{binaa} ‘without’
    \item \textit{boli} ‘because of’, ‘literally speaking’,  e.g. \textit{goli boli goTe pilaa thilaa} ‘there was a child called Goli’
    \item \textit{bhitare} ‘in’, ‘inside’
    \item \textit{majhire} ‘inside’, ‘in the midst of’
    \item \textit{laagi} ‘for’ e.g. \textit{raatidina laagi} ‘for day and night’
    \item \textit{sahite} ‘with’
\end{itemize}

\paragraph{Conjunctions}
Conjunction markers include \textit{o} / \textit{eban} / \textit{aau} ‘and’, \textit{kimbaa} / \textit{abaa}/\textit{athabaa} ‘or’, \textit{madhya} ‘also’, \textit{tathaapi} ‘still’, \textit{kintu} ‘but’ etc. 

\paragraph{Classifiers}
A classifier is a noun-related element but has no independent nominal reading.  Having insufficient referential or predicative content, it is not fully lexical.   -\textit{Taa} `one\textsubscript{[+def]}', \textit{Topaa} `drop', \textit{muThaa} `fist', \textit{gochhaa} `bundle',  \textit{jaNa} `one\textsubscript{[+Human]}', \textit{paTa} `slice', \textit{asaraa} `shower', \textit{menchaa}, etc.  are usually are identified as classifiers.  

\paragraph{Complex verbs}
Complex verb constructions like the combination of a verbal with a nominal (N-V sequences), and the combination of a verbal with a verbal (V-v sequences) are found in Odia. 

\paragraph{Serial verbs}
Odia is a verb serializing language.  A series of verbs along with their complements and adjuncts (if any) can occur in a single clause having a common subject.  Very often, the series of verbs have a common object too. 

\paragraph{Verbal Nouns}
Many verbal nouns are found in Odia, such as \textit{chaasa} ‘ploughing’, \textit{chaaDa} ‘release’, \textit{maajaNaa} ‘bath’, \textit{rahaNi} ‘stay’ \textit{bikaa} ‘selling’, \textit{baahuDaa} ‘return’.  Some verbal nouns have been borrowed from Sanskrit, e.g. \textit{anubhaba} ‘feeling’, \textit{bidroha} ‘revolution’, \textit{prabesha} ‘entrance’, \textit{sthiti} ‘existence’, etc. which are used along with a light verb in Odia. 

\paragraph{Copular sentences}
Copular constructions are usually sentences with a subject and a predicate.  The predicate may be either a noun (nominal predicate) or an adjective (adjectival predicate).

\paragraph{Adverbs}

Like English, Odia also has Time, Place, and Manner adverbials.

\paragraph{Finite Verbal Forms}
Agreement features contribute to the finiteness of a verbal form in Odia. All the finite verbal forms have an agreement in concord with the subject NP.  The agreement features of the verbal form are marked for the person, number, and honorific of the subject NP.  

\paragraph{The Infinitive}
In Odia, the infinitival form is realized by the verbal ending – \textit{ibaaku} ‘to do’.

\paragraph{The Conditional affix \textit{-ile} (or \textit{–le})}
The morpheme \textit{-ile} (or \textit{–le}) functions as a conditional marker.  It is suffixed to the bare verbal root.  It is nonfinite as it does not carry any Agr feature and thus can co-occur in a verbal form irrespective of person, number, or gender of the subject.

\begin{figure*}[!ht]
\begin{center}
    \includegraphics[scale=0.63]{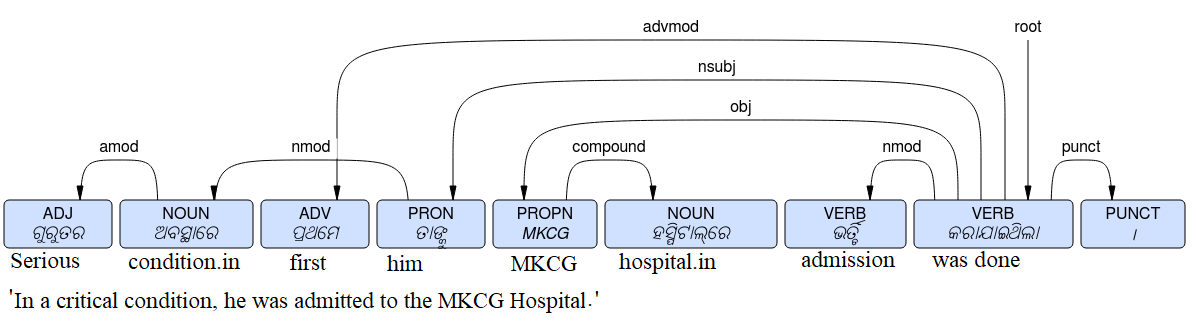}
    \caption{Passivization with Adverbial modifier construction in UD Odia}
    \label{fig:tree1}
    \end{center}
\end{figure*}

\begin{figure}[!ht]
\begin{center}
\includegraphics[scale=0.65]{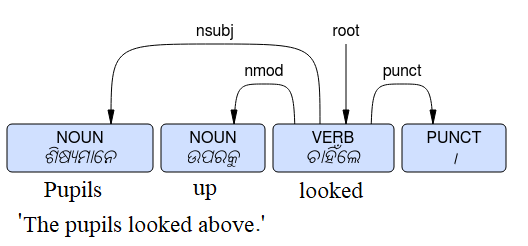}
\caption{Finite intransitive verb construction in UD Odia}
\label{fig:tree2}
\end{center}
\end{figure}

\begin{figure*}[ht]
\begin{center}
   \includegraphics[scale=0.67]{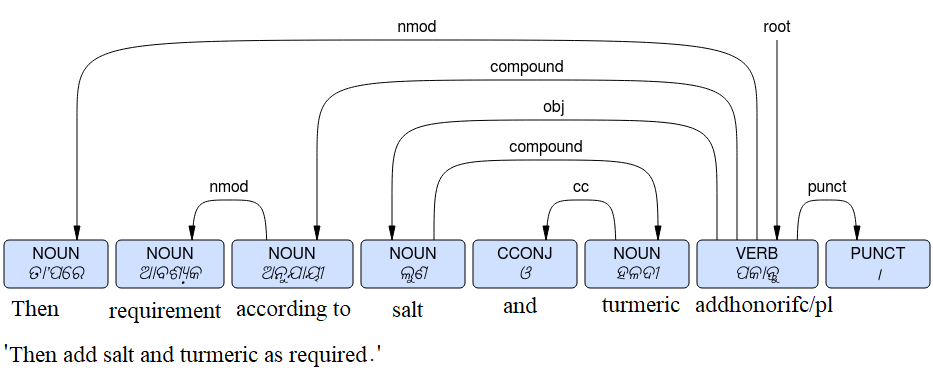}
    \caption{Finite verb in imperative sentence in UD Odia}
    \label{fig:tree3}
    \end{center}
\end{figure*}

\begin{figure*}[!ht]
\begin{center}
   \includegraphics[scale=0.67]{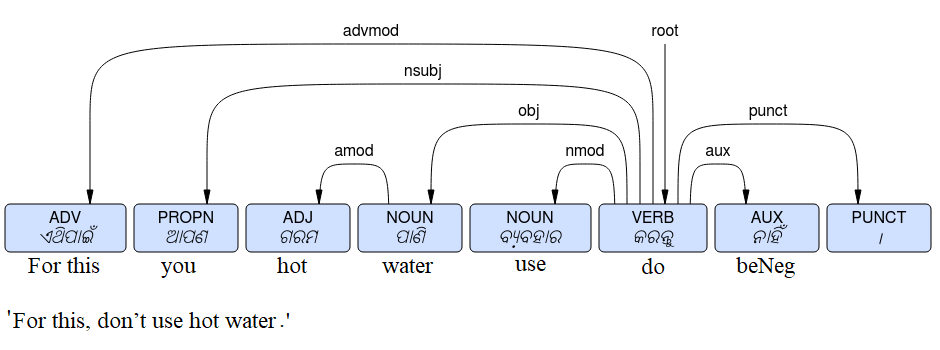}
    \caption{Main verb construction in UD Odia}
    \label{fig:tree5}
    \end{center}
\end{figure*}

\begin{figure*}[!ht]
\begin{center}
   \includegraphics[scale=0.67]{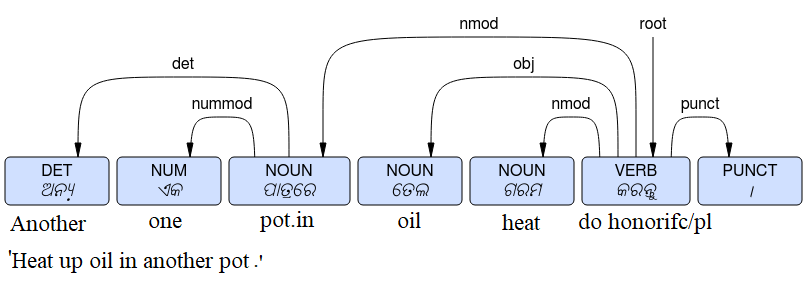}
    \caption{Finite verb with Noun modifier construction in UD Odia}
    \label{fig:tree4}
    \end{center}
\end{figure*}

\section{Related Work}
\label{sect:related_work}
Under the leadership of IIIT-Hyderabad, a consortium was formed in 2013 to start a project sponsored by TDIL (Government of India), called Development of Dependency Treebank for Indian Languages.\footnote{\url{http://meity.gov.in/content/language-computing-group-vi}} This project aimed to restore annotation work in monolingual treebanks for various languages such as Hindi, Marathi, Bengali, Kannada, and Malayalam. To achieve this model, the P{\=a}{\d n}inian K{\=a}raka Dependency scheme was followed~\cite{begum2008dependency,husain2010icon,bhat2017exploiting,ojha2020universal}. The same annotation scheme was used to annotate data in Telugu, Urdu, and Kashmiri. 

NLP research of Odia has led to development of a statistical POS tagger~\cite{ojha2015training}, neural network based POS tagger ~\cite{das2014novel}, POS tagging using Support Vector Machine (SVM) \cite{das2015part},  a shallow parsing tool \footnote{\url{http://calts.uohyd.ac.in/calts/sptil-parser.html}}, and English-Odia machine translation system~\cite{odiencorp1}.

Within the Universal Dependencies framework, as of UD release 2.8, treebanks
and parsers are available for Bhojpuri, Hindi, Marathi, Sanskrit,  Tamil,
Telugu and Urdu~\cite{11234/1-3687}.
Nevertheless, there is no
prior work on Odia dependency treebanking and parser.

\section{Data and Methodology}
\label{sect:methodology}

To collect Odia text, we used \textit{Samanantar}, the largest parallel corpora collection for 11 Indian languages \cite{ramesh2021samanantar}. The parallel corpora collection includes English-Odia parallel text that covers many domains. We selected the Odia sentences of word length between 5 to 15 words per sentence. For annotation, all selected sentences are converted into CoNLL-U format consisting of 10 fields \cite{buchholz2006conll}. The fields are ``ID", ``Word", ``Lemma", ``UPOS", ``XPOS", ``FEATS", ``HEAD", ``DEPREL", ``DEPS", and ``MISC". The ``UPOS" tags are based on the universal POS tags \footnote{\url{https://universaldependencies.org/u/pos/}} following the UD guidelines, version 2. For ``XPOS", we annotated according to Bureau of Indian Standards (BIS) Part of Speech (POS) tagset \footnote{\url{http://tdil-dc.in/tdildcMain/articles/134692Draft\%20POS\%20Tag\%20standard.pdf}} guideline released by the department of information technology ministry of communications \& information technology, the government of India. The guideline includes a POS tagset for the Odia language. The dependency relations were marked on Universal dependency tags which is an updated version of Stanford Dependencies~\cite{de-marneffe-etal-2014-universal}. Out 17 UPOS tags, we use 15 UPOS tag in this dataset, while out of 37 dependency tags, we use only 24 tags (see the Table~\ref{tab:upos} \& \ref{table:odiadeprel}).
The annotation task was performed by 6 native Odia speakers including 2 linguists.
\begin{table}[h!]
\begin{center}

\small\addtolength{\tabcolsep}{-2pt}
\begin{tabular}{|c|c|c|c|c}
    \hline
    \textbf{UPOS Tags} & \textbf{UPOS description } & \textbf{Statistics}\\
      \hline
      NOUN & Noun & 570\\
      \hline
      VERB & Verb & 234\\
      \hline
      PUNCT & Punctuation & 192\\
      \hline
      PROPN & Proper noun & 170\\
    \hline
    ADJ & Adjective & 102\\
    \hline
    ADP & Adposition & 82\\
    \hline
    DET & Determiner & 75\\
    \hline
    PRON & Pronoun & 55\\
    \hline
    CCONJ & Coordinating conjunction & 48\\
    \hline
     ADV & Adverb & 45\\
     \hline
     NUM & Numeral & 26\\
     \hline
      PART & Particle & 23\\
      \hline
    AUX & Auxiliary & 13\\
    \hline
    SCONJ & Subordinating conjunction & 7\\
    \hline
    SYM & Symbol & 1\\
    \hline
\end{tabular}
\caption{Statistics of used UPOS Tags in the Odia treebank} 
\label{tab:upos}
 \end{center}
\end{table}
\begin{table}[!h]
  \begin{center}
    \small\addtolength{\tabcolsep}{-5pt}
    \begin{tabular}{|c|c|c|}
    \hline
    \textbf{UD Relations} & \textbf{Description} &\textbf{Statistics}\\
    \hline
    advmod & Adverbial modifier & 67\\\hline\hline
    advmod & Locative adverbial modifier & 4\\\hline\hline
    amod & Adjectival modifier of noun & 109\\\hline\hline
    aux & Auxiliary verb & 9\\\hline\hline
    case & Case marker & 1\\\hline\hline
    cc & Coordinating conjunction & 1\\\hline\hline
    ccomp & Clausal complement & 2\\\hline\hline
    compound & Compound & 85\\\hline\hline
    conj & Non-first conjunct & 2\\\hline\hline
    cop & Copula & 1 \\\hline\hline
    det & Determiner & 72\\\hline\hline
    fixed & Non-first word of fixed expression & 63\\\hline\hline
    flat & non-first word of flat structure & 51\\\hline\hline
    goeswith & Non-first part of broken word & 6\\\hline\hline
    iobj & Indirect object & 72\\\hline\hline
    mark & Subordinating marker & 52\\\hline\hline
    nmod & Nominal modifier of noun & 287\\\hline\hline
    nsubj & Nominal subject & 136\\\hline\hline
    nummod & Numeric modifier & 33\\\hline\hline
    obj & Direct object & 122\\\hline\hline
    obl & Oblique nominal & 1\\\hline\hline
    punct & Punctuation & 192\\\hline\hline
    root & Root & 174\\\hline\hline
    xcomp & Open clausal complement & 1\\\hline
  \end{tabular}
  \caption{UD relations used in Odia trebank}
  \label{table:odiadeprel}
  \end{center}
\end{table}

\section{Experiment and Results}
\label{sect:experiment}
As mentioned earlier, the Odia treebank was manually annotated using the UD annotation framework. In this, we have built Odia parser on 2026 tokens using the UDPipe open-source tool~\cite{straka2017tokenizing}. We conducted our experiment in two parts. The first experiment was conducted on 50 sentences, while the second experiment was conducted on the rest of the dataset. We used a cross-validation 90:10 average for the data splitting where the batch size, learning rate, and dropout were 50, 0.005, and 0.10, respectively; while the other hyperparameters were randomized. The results are demonstrated in Table~\ref{tab:odiaparser}: 
\begin{table}[!h]
  \begin{center}
    \small\addtolength{\tabcolsep}{-1pt}
    \begin{tabular}{|c|c|c|c|c|}
      \hline
    \textbf{Tokenization}  & \textbf{UPOS} & \textbf{XPOS} & \textbf{UAS} & \textbf{LAS} \\
      \hline
      81.82\% & 48.25\% & 45.0\% & 36.62\% & 16.91\% \\\hline
      86.6\% & 64.1\% & 63.78\% & 42.04\% & 21.34\% \\\hline
    \end{tabular}
    \caption{Results of Odia Parser}
    \label{tab:odiaparser}
  \end{center}
\end{table}
Due to the small size of the data, the parser's accuracy is very low except on Tokenization. 
\section{Linguistic Analysis}
\label{sect:analysis}

We are providing few sample tree constructions along with their linguistics analysis in ~\Cref{fig:tree1,fig:tree2,fig:tree3,fig:tree4,fig:tree5}

In~\Cref{fig:tree1}, \textit{‘karaajaaithilaa’} is a finite verb. So, it forms the root. The adjective \textit{‘guruttara’} modifies the noun \textit{‘abasthaare’}. The main verb has \textit{‘taanku’} as the external argument (the subject) and \textit{‘MKCG hospitalre’} as internal argument (the object) of it. It has the adverbial modifier \textit{‘prathame’}.

In ~\Cref{fig:tree2}, the finite intransitive verb \textit{‘chaahinle’} ‘wanted’ is the root of the sentence. It takes \textit{‘shishyamaane’} ‘pupils’ as the subject argument. Being intransitive, it does not take any object or internal argument.

In ~\Cref{fig:tree3}, \textit{‘pakaantu’} ‘put’ is the finite verb, which forms the root. Being an imperative sentence, the subject noun is not realized, and \textit{‘luNa o haLadi’} ‘salt and turmeric’ functions as the object of the sentence.  \textit{‘taa pare’} ‘after that’ functions as the adverbial modifier. \textit{‘aabashyaka anujaayi’} ‘as per the requirement’ functions as a nominal modifier for \textit{‘luNa o haLadi’} ‘salt and turmeric’.

In ~\Cref{fig:tree5}, the main verb \textit{‘karantu’} ‘do Pl\slash Honorific ‘ takes \textit{‘aapaNa’} ‘you’ as the subject noun and \textit{‘paaNi’} ‘water’ as the object. 
It is a negative sentence, and the negative auxiliary \textit{‘naahin’} ‘be-Neg’ occurs at the end of the sentence. The object \textit{‘paaNi’} ‘water’ is modified by the adjective \textit{‘garama’} ‘hot’. \textit{‘ethipaain’} ‘because of this’ functions as the adverbial modifier for the sentence.

In ~\Cref{fig:tree4}, \textit{‘karantu’} ‘do Pl/Honorific ‘ which is a finite verb, forms the root of the sentence. the det \textit{‘anya’} ‘another’ and the numeral modifier \textit{‘eka’}’one’ modify the locative modifier \textit{‘paatrare’} ‘in a pot’. The nsubject \textit{‘aapaNa’} ‘you Pl/Honorific ‘ is not realized in the sentence. The main verb \textit{‘karantu’} ‘do Pl/Honorific ‘ takes \textit{‘tela’} ‘oil’ as the object and  \textit{‘paatrare’} ‘in a pot’  as the locative modifier.





\section{Conclusion and Future Work}
\label{sect:conclusion}
We presented the first UD Odia treebank aimed for linguistic research and applications in NLP, primarily for POS tagging, parser, semantic analyzer, and machine translation. Also, we built a preliminary Odia parser using the UDPipe tool. The accuracy of the Odia parser is 86.6\% Tokenization, 64.1\% UPOS, 63.78\% XPOS, 42.04\% UAS and 21.34\% LAS.

Future research direction includes: \textit{i)}  enrich the Odia treebank with more annotated data for training, development, and validation, \textit{ii)} including lemma for the Odia tokens, \textit{iii)} perform detail morphological analysis, and \textit{iv)} experiment with neural network based models for performance evaluation.  
\\

\section{Acknowledgements}
\label{sect:acknowledgeent}

Atul Kr. Ojha would like to acknowledge the EU’s Horizon 2020 Research and Innovation programme through the ELEXIS project under grant agreement No. 731015.

\section{References}\label{reference}

\bibliographystyle{lrec2022-bib}
\bibliography{biblio}

\begin{thebibliography}{}

\bibitem[\protect\citename{Begum \bgroup et al.\egroup
  }2008]{begum2008dependency}
Begum, R., Husain, S., Dhwaj, A., Sharma, D.~M., Bai, L., and Sangal, R.
\newblock (2008).
\newblock Dependency annotation scheme for {I}ndian languages.
\newblock In {\em Proceedings of the Third International Joint Conference on
  Natural Language Processing: Volume-II}.

\bibitem[\protect\citename{Bhat}2017]{bhat2017exploiting}
Bhat, R.~A.
\newblock (2017).
\newblock {\em Exploiting linguistic knowledge to address representation and
  sparsity issues in dependency parsing of {I}ndian languages}.
\newblock {Ph.D.} thesis, PhD thesis, International Institute of Information
  Technology, India.

\bibitem[\protect\citename{Buchholz and Marsi}2006]{buchholz2006conll}
Buchholz, S. and Marsi, E.
\newblock (2006).
\newblock Conll-x shared task on multilingual dependency parsing.
\newblock In {\em Proceedings of the tenth conference on computational natural
  language learning (CoNLL-X)}, pages 149--164.

\bibitem[\protect\citename{Das and Patnaik}2014]{das2014novel}
Das, B.~R. and Patnaik, S.
\newblock (2014).
\newblock A novel approach for odia part of speech tagging using artificial
  neural network.
\newblock In {\em Proceedings of the International Conference on Frontiers of
  Intelligent Computing: Theory and Applications (FICTA) 2013}, pages 147--154.
  Springer.

\bibitem[\protect\citename{Das \bgroup et al.\egroup }2015]{das2015part}
Das, B.~R., Sahoo, S., Panda, C.~S., and Patnaik, S.
\newblock (2015).
\newblock Part of speech tagging in odia using support vector machine.
\newblock {\em Procedia Computer Science}, 48:507--512.

\bibitem[\protect\citename{de Marneffe \bgroup et al.\egroup
  }2014]{de-marneffe-etal-2014-universal}
de~Marneffe, M.-C., Dozat, T., Silveira, N., Haverinen, K., Ginter, F., Nivre,
  J., and Manning, C.~D.
\newblock (2014).
\newblock Universal {S}tanford dependencies: A cross-linguistic typology.
\newblock In {\em Proceedings of the Ninth International Conference on Language
  Resources and Evaluation ({LREC}'14)}, pages 4585--4592, Reykjavik, Iceland,
  May. European Language Resources Association (ELRA).

\bibitem[\protect\citename{Husain \bgroup et al.\egroup }2010]{husain2010icon}
Husain, S., Mannem, P., Ambati, B.~R., and Gadde, P.
\newblock (2010).
\newblock The {ICON}-2010 tools contest on {Indian} language dependency
  parsing.
\newblock {\em Proceedings of ICON-2010 Tools Contest on Indian Language
  Dependency Parsing, ICON}, 10:1--8.

\bibitem[\protect\citename{Ojha and Zeman}2020]{ojha2020universal}
Ojha, A.~K. and Zeman, D.
\newblock (2020).
\newblock Universal {Dependency Treebanks for Low-Resource Indian Languages:
  The Case of Bhojpuri}.
\newblock In {\em Proceedings of the WILDRE5--5th Workshop on Indian Language
  Data: Resources and Evaluation}, pages 33--38.

\bibitem[\protect\citename{Ojha \bgroup et al.\egroup }2015]{ojha2015training}
Ojha, A.~K., Behera, P., Singh, S., and Jha, G.~N.
\newblock (2015).
\newblock Training \& evaluation of pos taggers in indo-aryan languages: a case
  of hindi, odia and bhojpuri.
\newblock In {\em the proceedings of 7th Language \& Technology Conference:
  Human Language Technologies as a Challenge for Computer Science and
  Linguistics}, pages 524--529.

\bibitem[\protect\citename{Parida \bgroup et al.\egroup }2020a]{odiencorp1}
Parida, S., Bojar, O., and Dash, S.~R.
\newblock (2020a).
\newblock Odiencorp: Odia--english and odia-only corpus for machine
  translation.
\newblock In {\em Smart Intelligent Computing and Applications}, pages
  495--504. Springer.

\bibitem[\protect\citename{Parida \bgroup et al.\egroup }2020b]{odiencorp2}
Parida, S., Dash, S.~R., Bojar, O., Motlicek, P., Pattnaik, P., and Mallick,
  D.~K.
\newblock (2020b).
\newblock {O}di{E}n{C}orp 2.0: {O}dia-{E}nglish parallel corpus for machine
  translation.
\newblock In {\em Proceedings of the WILDRE5{--} 5th Workshop on Indian
  Language Data: Resources and Evaluation}, pages 14--19, Marseille, France,
  May. European Language Resources Association (ELRA).

\bibitem[\protect\citename{Ramesh \bgroup et al.\egroup
  }2021]{ramesh2021samanantar}
Ramesh, G., Doddapaneni, S., Bheemaraj, A., Jobanputra, M., AK, R., Sharma, A.,
  Sahoo, S., Diddee, H., Kakwani, D., Kumar, N., et~al.
\newblock (2021).
\newblock Samanantar: The largest publicly available parallel corpora
  collection for 11 indic languages.
\newblock {\em arXiv preprint arXiv:2104.05596}.

\bibitem[\protect\citename{Sahoo}2001]{sahoo2001oriya}
Sahoo, K.
\newblock (2001).
\newblock {\em Oriya verb morphology and complex verb constructions}.
\newblock NTNU Trondheim.

\bibitem[\protect\citename{Straka and Strakov{\'a}}2017]{straka2017tokenizing}
Straka, M. and Strakov{\'a}, J.
\newblock (2017).
\newblock Tokenizing, {POS} tagging, lemmatizing and parsing {UD} 2.0 with
  {UDPipe}.
\newblock In {\em Proceedings of the CoNLL 2017 Shared Task: Multilingual
  Parsing from Raw Text to Universal Dependencies}, pages 88--99.

\bibitem[\protect\citename{Zeman and et al.}2021]{11234/1-3687}
Zeman, D. and et~al.
\newblock (2021).
\newblock Universal dependencies 2.8.1.
\newblock {LINDAT}/{CLARIAH}-{CZ} digital library at the Institute of Formal
  and Applied Linguistics ({{\'U}FAL}), Faculty of Mathematics and Physics,
  Charles University.

\end{thebibliography}

\end{document}